\documentclass{article} 
\usepackage{iclr2025_conference,times}
\usepackage[utf8]{inputenc}
\usepackage{wrapfig}
\usepackage{hyperref}
\usepackage{graphicx}
\usepackage{url}
\usepackage{listings}
\usepackage[most]{tcolorbox}
\usepackage{tikz}

\definecolor{main}{HTML}{0C8841}    
\definecolor{sub}{HTML}{FFFDEF}
\definecolor{black}{HTML}{000000}

\lstdefinestyle{mystyle}{
    basicstyle=\ttfamily, 
    breaklines=true, 
    columns=fullflexible, 
}
\newtcolorbox{promptbox}[2][]{%
lower separated=false,
colback=black!5, 
colframe=black!80, 
fonttitle=\bfseries, 
title=#2, 
boxrule=0.5mm, 
width=\textwidth, 
before=\vspace{10pt}, 
after=\vspace{10pt}, 
#1,
arc=5mm, 
enhanced,
breakable 
}

\usepackage{amsmath,amsfonts,bm}

\def\eqref#1{equation~\ref{#1}}

\def\1{\bm{1}}

\DeclareMathAlphabet{\mathsfit}{\encodingdefault}{\sfdefault}{m}{sl}
\SetMathAlphabet{\mathsfit}{bold}{\encodingdefault}{\sfdefault}{bx}{n}

\title{InfoQuest: Evaluating Multi-Turn Dialogue Agents for Open-Ended Conversations with Hidden Context}

\author{Bryan L. M. de Oliveira$^{1,2,*}$, Luana G. B. Martins$^{1,2}$, Bruno Brandão$^{1,2}$, Luckeciano C. Melo$^{1,3}$ \\
$^1$Advanced Knowledge Center for Immersive Technologies -- AKCIT, Brazil\\
$^2$Institute of Informatics, Federal University of Goiás, Brazil\\
$^3$OATML, University of Oxford, United Kingdom\\
\texttt{*bryanlincoln@discente.ufg.br}
}

\iclrfinalcopy 
\begin{document}

\maketitle

\begin{abstract}
Large language models excel at following explicit instructions, but they often struggle with ambiguous or incomplete user requests, defaulting to verbose, generic responses instead of seeking clarification. We introduce InfoQuest, a multi-turn chat benchmark designed to evaluate how dialogue agents handle hidden context in open-ended user requests. This benchmark presents intentionally ambiguous scenarios that require models to engage in information-seeking dialogue by asking clarifying questions before providing appropriate responses. Our evaluation of both open and closed models reveals that, while proprietary models generally perform better, all current assistants struggle to gather critical information effectively. They often require multiple turns to infer user intent and frequently default to generic responses without proper clarification. We provide a systematic methodology for generating diverse scenarios and evaluating models' information-seeking capabilities, which can be leveraged to automatically generate data for self-improvement. We also offer insights into the current limitations of language models in handling ambiguous requests through multi-turn interactions.
\end{abstract}

\section{Introduction}


Large language models (LLMs) have demonstrated remarkable capabilities in following explicit instructions and engaging in task-oriented dialogue. However, when users provide incomplete or ambiguous requests -- a common occurrence in real-world interactions -- these models often default to producing verbose, generic responses rather than seeking clarification \citep{kuhn2022clam,kim2023tree,chi2024clarinet,zhou2024archer,rahmani2023survey}. The ability to ask clarification questions is crucial for conversational systems to understand a user's underlying needs, especially with limited input \citep{rahmani2023survey}. Furthermore, in order to maximize the expected quality of a conversation, dialogue agents should reason about the stochastic transitions within a conversation to select the optimal response at each turn \citep{chen2025broaden}.

\begin{figure}[t]
   \includegraphics[width=\linewidth]{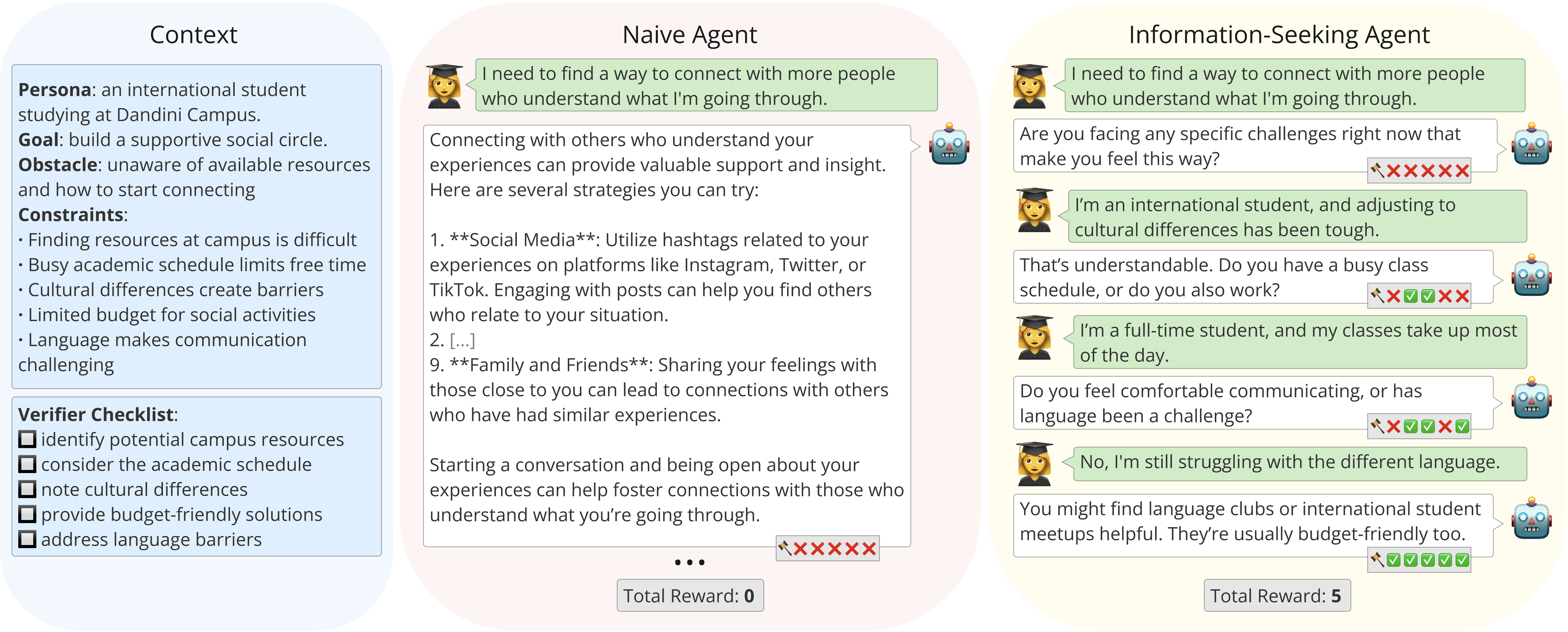}
   \caption{\textbf{Naive vs. information-seeking agents handling ambiguous user requests}. Left: context. Center: naive agent's verbose response. Right: information-seeking agent's targeted questions.}
   \label{fig:example}
\end{figure}

As a motivating example, consider the conversation shown in Figure~\ref{fig:example}. The left block provides the conversation context. The center block illustrates how a naive agent responds with a lengthy, generic explanation, making assumptions about the user's context. In contrast, the right block shows an information-seeking agent engaging in focused dialogue, asking targeted clarifying questions to understand the specific scenario before providing a tailored response. This balanced approach of gathering key information while maintaining conversation flow is crucial -- excessive questioning can frustrate users, while insufficient context-gathering leads to unhelpful generic responses. The information-seeking behavior enables agents to provide more accurate, personalized assistance by understanding the user's unique situation, constraints, and goals. However, current LLMs often fail to strike this balance, defaulting to overly broad responses that may waste the user's time without addressing their actual needs. Effective information-seeking requires carefully weighing when additional context is truly needed versus when enough information exists to provide meaningful help.

Evaluating dialogue agents in open-ended conversations presents unique challenges. In constrained environments like games or structured tasks, quantifying how well a model collects relevant context is relatively straightforward \citep{hausknecht2020detective,yao2022webshop,abdulhai2023lmrl,zhou2024archer}. However, measuring this capability becomes significantly more complex in open-ended, multi-turn settings that better reflect everyday human-AI interactions. The key challenge lies in the unbounded nature of these interactions; there is no predefined "correct" sequence of questions or complete set of information to gather, making it difficult to establish ground truth or quantitative metrics. While existing benchmarks have focused on structured tasks or specific dialogue abilities \citep{bai2024mtbench101,kwan2024mteval}, an evaluation framework that can assess how models handle the inherent ambiguity and contextual complexity of natural conversations remains necessary.


To address this gap, we introduce \textit{InfoQuest}, a multi-turn chat benchmark designed to evaluate how dialogue agents handle hidden context in user requests. It simulates conversational settings with intentionally open-ended and ambiguous requests by generating seed messages that could plausibly come from multiple distinct personas, each with their own goals and constraints. This controlled ambiguity requires agents to demonstrate effective information-seeking behavior through clarifying questions and progressively infer the user's specific context before providing a satisfactory answer. The benchmark uses a combination of LLMs to simulate realistic user responses. We leverage other LLMs for role-playing the user in our dynamic scenarios \citep{zhou2024sotopia} and evaluate the assistant's ability to gather critical information through targeted questioning.


Our evaluation reveals that, while proprietary models generally outperform open models, all current assistants struggle to effectively handle hidden information across diverse user scenarios. Notably, we find that models require multiple turns to infer user intent and address latent requests, demonstrating poor turn efficiency. Qualitatively, we observe that models often default to generic responses without asking clarification questions, highlighting a key area for improvement in developing more interactive and context-aware conversational agents.

\textbf{Contributions.} This work contributes to the field by introducing InfoQuest, a novel benchmark for evaluating how language models handle open-ended conversations with hidden context. We develop a systematic methodology for generating diverse, ambiguous scenarios that require models to engage in multi-turn dialogue to uncover critical information. Furthermore, the data generated through InfoQuest's automated pipeline can be used for self-improvement, enabling models to learn from their interactions and enhance their ability to handle ambiguous queries. Additionally, we conduct comprehensive experiments comparing both open and proprietary models on their ability to handle ambiguous requests through multi-turn interactions, providing insights into the current limitations of LLMs in this area and identifying key challenges in information-seeking dialogue.

\section{Related Work}

Our work builds on prior research in clarification questions and multi-turn dialogue evaluation. Recent work has explored various approaches to handling ambiguous queries. \citet{kuhn2022clam} proposed a two-stage framework that first classifies whether a question is ambiguous and then generates appropriate clarifying questions. However, their evaluation focused on single-turn interactions using paired ambiguous and unambiguous questions. \citet{kim2023tree} developed a method to generate comprehensive trees of disambiguations to address ambiguous queries in a single response, in contrast to our focus on interactive, multi-turn clarification. \citet{chi2024clarinet} introduced an approach for selecting clarifying questions that maximize certainty in book search tasks, though their work was limited to this specific domain rather than open-ended dialogue.

Several benchmarks have been developed to evaluate multi-turn dialogue capabilities in constrained settings. \citet{hausknecht2020detective} created interactive fiction games that require information gathering through a text interface. \citet{yao2022webshop} developed a shopping website simulation where agents must navigate constraints to complete purchases. \citet{abdulhai2023lmrl} proposed environments for training and evaluating reinforcement learning with language models. While these works provide valuable insights into structured information gathering, they rely on rigid rules and predefined success criteria that may not reflect the complexity of open-ended dialogue.

Recent work has explored fine-grained evaluation of multi-turn dialogue abilities. \citet{bai2024mtbench101} used GPT-4 to generate evaluation conversations with fixed user messages, providing detailed assessment across multiple capabilities. \citet{kwan2024mteval} extended existing datasets by adding predefined follow-up questions to evaluate conversation progression. While these benchmarks offer systematic evaluation approaches, they differ from our work in that they focus on predefined dialogue flows rather than dynamic information gathering in response to ambiguous queries.

\section{InfoQuest}
\label{method}

\begin{figure}[t]
   \includegraphics[width=\linewidth]{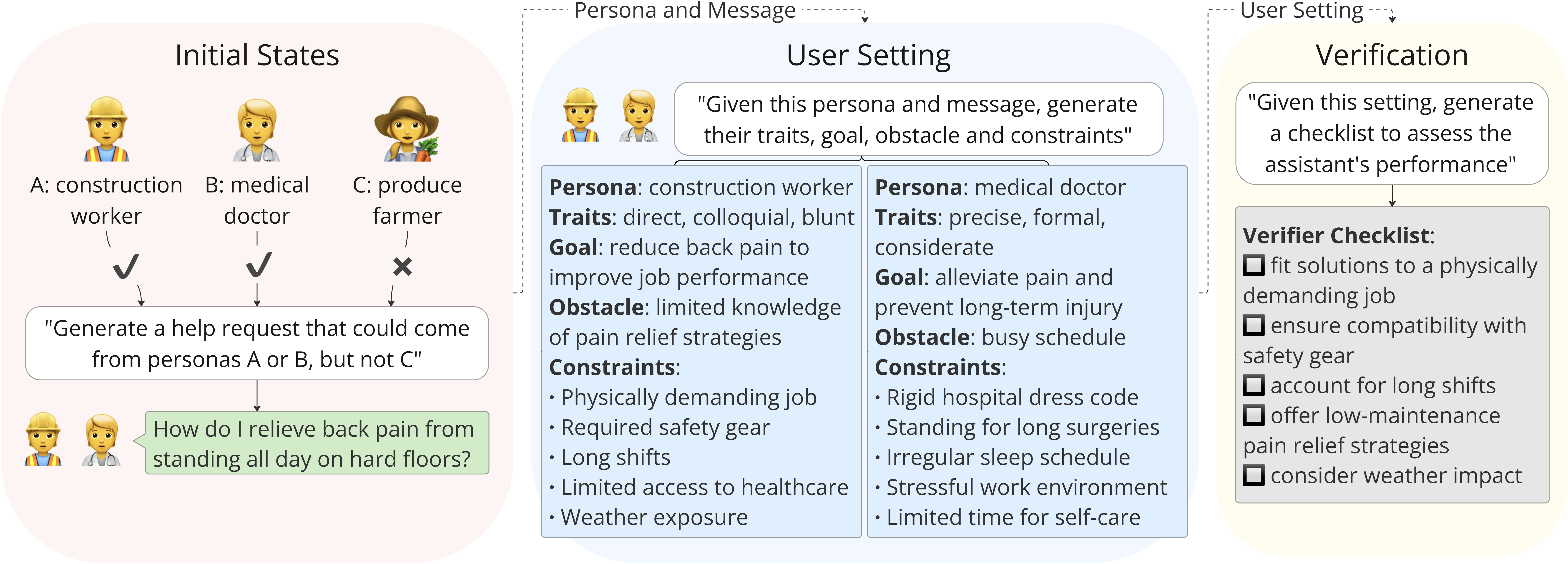}
   \caption{\textbf{InfoQuest's three-stage benchmark construction process.} Left: initial state generation by selecting personas and creating ambiguous messages. Center: user setting with persona traits, goals, obstacles and constraints. Right: generation of a checklist to evaluate information gathering.}
   \label{fig:diagram}
\end{figure}

InfoQuest is a benchmark designed to evaluate dialogue agents' ability to handle ambiguity in multi-turn conversations. It consists of three main components, as illustrated in Figure \ref{fig:diagram}: initial state generation, user simulation, and a verification process. These components work together to create a dynamic environment where the evaluated assistant must gather critical information through clarifying questions, simulating real-world interactions where user requests are often underspecified.

\subsection{Components of InfoQuest}

\textbf{Initial State Distribution.} To generate evaluation scenarios, we create ambiguous initial messages that could plausibly come from multiple distinct personas sourced from the PersonaHub dataset \citep{ge2024personahub}, each with their own goals and constraints. For each scenario, we select three personas (A, B, and C) and generate a seed message that could originate from either A or B, but not C. This controlled ambiguity ensures that the assistant must ask clarifying questions to determine the specific context while maintaining enough direction for meaningful interaction. Examples of personas are shown in Appendix~\ref{ap:personas}, and generated seed messages are shown in Appendix~\ref{ap:infoquest_examples}.

\textbf{User Simulation.} For each scenario, personas are augmented with three distinct personality traits that influence their communication style and response patterns. This setup ensures that the assistant must adapt its information-seeking strategy to the user's personality, enhancing the realism and complexity of the interaction. We also generate a comprehensive "setting" that defines the evaluation scenario, which includes:

\begin{itemize}
   \item A description of the user's goal, its importance, and key constraints to accomplish it;
   \item A specific, non-trivial goal that aligns with the persona's context and naturally prompts them to seek assistance;
   \item A realistic obstacle or challenge that requires the assistant to ask clarifying questions to understand the full context;
   \item Five key constraints that combine specific factors with their complications (e.g., "budget constraints limit equipment options").
\end{itemize}

The description, goal, and obstacle components provide the user simulator with additional context to better understand and engage with the setting, while the constraints represent the specific information that assistants need to discover through questioning. The simulator maintains consistency with the selected persona while revealing information gradually based on the assistant's questions.

\textbf{Verification Process.} For each scenario, we generate a checklist of five specific yes/no questions that evaluate how well the assistant gathers critical information about the user's goals, constraints, and obstacles. These questions are designed to be easily verifiable while covering key aspects of the user's context that must be uncovered through dialogue. In preliminary experiments, this yes/no setup provided the most consistent results, since they are easily and objectively verifiable, while also helping to mitigate the inherent biases that judge models may exhibit. A judge model evaluates the progress of each conversation by assessing the checklist questions after every assistant message. For each question, the judge considers the user context (persona, description, goal, and obstacle) along with the most recent user and assistant messages to make a binary yes/no assessment.

\subsection{Simulation and Evaluation Methodology}

The interaction process begins with an ambiguous prompt and proceeds for up to 10 turns between the user simulator and the assistant. We instruct the user simulator to reveal at most one piece of information per message, compelling the assistant to ask a sequence of targeted questions. The evaluation results update the user model with information about pending objectives, ensuring the conversation remains focused on uncovering all necessary information. Responses remain intentionally vague when the assistant makes progress, with subtle guidance provided after two unproductive turns. Conversations continue until all checklist items are satisfied or the maximum turn limit is reached. This methodology rewards effective information-seeking strategies and turn efficiency while penalizing overly broad or unfocused approaches.

\subsection{Operationalization and Implementation}

We designed InfoQuest as a sequential decision-making problem similar to a Partially Observable Markov Decision Process (POMDP), where each chat represents an episode, the chat history constitutes the state, the seed message defines the initial state, messages are actions, progress on the checklist provides the reward signal, user behavior determines the environment dynamics, and terminal states are reached upon satisfying all checklist items.

In our implementation, we selected Gemini 2.0 Flash~\citep{pichai2024gemini2} as the user simulator because it offered the best balance of cost-effectiveness and capability to reliably follow our complex role-playing instructions while maintaining consistent response patterns. To mitigate the risk of evaluation biases, we use a third model as judge, Selene 1 Mini \citep{alexandru2025selene}, an open general-purpose evaluator chosen for its reported performance and reliable evaluation capabilities. While we initially experimented with other small open models for these roles, we found they struggled with maintaining consistency across complex scenarios. Future work could explore fine-tuning approaches to enable the use of lighter-weight models for these components.

Our prompt design process followed an iterative development approach, involving multiple rounds of testing and refinement to ensure effectiveness across diverse scenarios. We initially validated the prompts on a small test set, allowing for rapid feedback and adjustments to optimize their ability to elicit desired responses from dialogue agents. This iterative process was crucial for maintaining consistency across different scenarios, ensuring fair and controlled evaluation of agent performance. The precise prompts used for both the user simulator and judge are provided in Appendix~\ref{ap:prompts}.

For each setting, we simulate conversations for up to 10 turns (20 messages total, alternating between assistant and user). This length balances the need for meaningful interaction depth with computational efficiency. Our publicly released dataset\footnote{Available at \url{https://huggingface.co/datasets/bryanlincoln/infoquest}} comprises 500 unique seed messages, each paired with two distinct settings for a total of 1,000 evaluation scenarios. Each scenario includes comprehensive metadata: the associated persona and traits, detailed setting information, and evaluation checklists. To facilitate reproducibility and comparative analysis, we also provide the complete conversation logs from all baseline models along with their corresponding verification results.

\begin{figure}[t]
   \includegraphics[width=\linewidth]{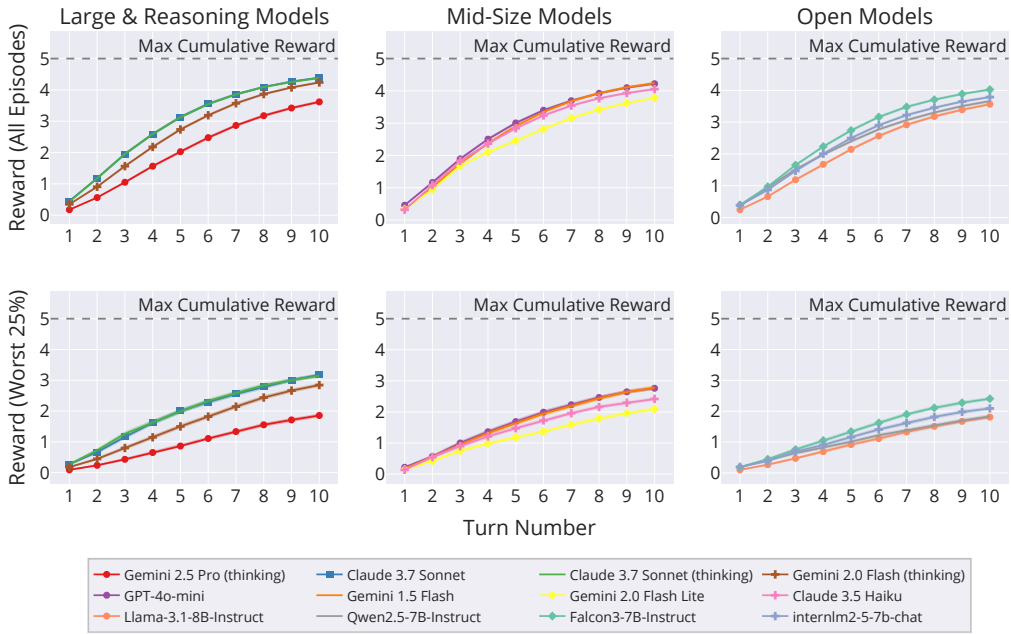}
   \caption{\textbf{Average cumulative reward of diverse dialogue agents on InfoQuest}. Top row: performance across all episodes. Bottom row: performance on the worst 25\% of episodes, where performance gaps become more apparent. Models are grouped by category: large \& reasoning models (left), mid-sized proprietary models (center), and open models (right). While all methods achieve non-trivial performance, there remains significant room for improvement in handling hidden context in open-ended conversations.}
   \label{fig:rewards_full}
\end{figure}

\section{Experiments and Discussion}
\label{results}

In this section, we evaluate several LLMs as assistants in the InfoQuest benchmark. Our aim is twofold: first, to assess the performance of existing models on our proposed benchmark, and second, to validate the effectiveness of the benchmark in differentiating model capabilities and highlighting areas for improvement in multi-turn dialogue with hidden context.

\textbf{Baselines.} We consider a wide range of mid-sized and large language models, both proprietary and open-weight, categorized by their size and reasoning capabilities. All models are ``Instruct'' versions optimized for chat capabilities, and we evaluate them in a zero-shot manner with three independent runs, reporting means with 95\% confidence intervals.

\textbf{Proprietary Models by Size.} For larger proprietary models, we evaluate \textbf{Claude 3.7 Sonnet} \citep{claude37modelcard} and \textbf{Gemini 2.5 Pro (thinking)} \citep{gemini25pro}. For mid-sized proprietary models, we include \textbf{Claude 3.5 Haiku} \citep{claude35modelcard}, \textbf{Gemini 1.5 Flash} \citep{team2024gemini15}, \textbf{Gemini 2.0 Flash Lite} \citep{pichai2024gemini2}, \textbf{Gemini 2.0 Flash} \citep{pichai2024gemini2}, and \textbf{GPT-4o-mini} \citep{hurst2024gpt4o}.

\textbf{Models with Enhanced Reasoning.} To assess the impact of advanced reasoning capabilities, we evaluate specialized versions of several models: \textbf{Claude 3.7 Sonnet (thinking)} \citep{claude37modelcard}, \textbf{Gemini 2.0 Flash (thinking)} \citep{pichai2024gemini2}, and \textbf{Gemini 2.5 Pro (thinking)} \citep{gemini25pro}. These variants incorporate additional reasoning steps and structured thinking processes in their responses.

\textbf{Open Models.} For open-weights models, we evaluate \textbf{Falcon3-7B-Instruct} \citep{falcon3}, \textbf{internlm2.5-7b-chat} \citep{cai2024internlm2}, \textbf{Qwen2.5-7B-Instruct} \citep{yang2024qwen2}, and \textbf{Llama-3.1-8B-Instruct} \citep{dubey2024llama}. These models were selected to ensure our results are reproducible in most academic settings.

\begin{table}[t]
   \centering
   \caption{\textbf{Ranking of proprietary (top) and open (bottom) LLMs by average reward and number of turns.} For returns, higher is better. For turns, lower is better. The maximum possible value is 5.0 for returns and 10 for turns. Lower turns indicate better information-seeking efficiency.}
   \label{tab:ranking}
   \begin{tabular}{lcc}
   \hline
   \textbf{Model Name} & \textbf{Return} & \textbf{Turns} \\
   \hline
   Claude 3.7 Sonnet & 4.39 ± 0.03 & 7.99 ± 0.08 \\
   Claude 3.7 Sonnet (thinking) & 4.38 ± 0.03 & 8.04 ± 0.08 \\
   Gemini 1.5 Flash & 4.24 ± 0.04 & 8.28 ± 0.08 \\
   Gemini 2.0 Flash (thinking) & 4.24 ± 0.04 & 8.47 ± 0.07 \\
   GPT-4o-mini & 4.22 ± 0.04 & 8.24 ± 0.08 \\
   Claude 3.5 Haiku & 4.05 ± 0.04 & 8.52 ± 0.08 \\
   Gemini 2.0 Flash Lite & 3.78 ± 0.04 & 8.96 ± 0.07 \\
   Gemini 2.5 Pro (thinking) & 3.63 ± 0.05 & 9.24 ± 0.06 \\
   \hline
   Falcon3-7B-Instruct & 4.03 ± 0.04 & 8.60 ± 0.08 \\
   InternLM2.5-7b-chat & 3.79 ± 0.04 & 8.89 ± 0.07 \\
   Qwen2.5-7B-Instruct & 3.66 ± 0.05 & 8.98 ± 0.07 \\
   Llama-3.1-8B-Instruct & 3.56 ± 0.05 & 9.20 ± 0.06 \\
   \hline
   \end{tabular}
\end{table}

We highlight and analyze the following research questions:

\textbf{How does InfoQuest distinguish the chat capabilities under hidden context of current dialogue agents?} To answer this question, we directly analyze the average cumulative reward obtained by the baselines over the multi-turn interaction, as presented in Figure \ref{fig:rewards_full} (top row) and Table \ref{tab:ranking}. Claude 3.7 Sonnet emerges as the clear leader in both return and turn efficiency, with its standard version slightly outperforming the thinking variant. Gemini 1.5 Flash, Gemini 2.0 Flash (thinking), and GPT-4o-mini form the next performance tier, with very similar results. Interestingly, Gemini 2.0 Flash (thinking) falls slightly short of Gemini 1.5 Flash's performance, suggesting that newer models with enhanced reasoning capabilities do not necessarily outperform older models in information-seeking tasks. Claude 3.5 Haiku follows in the next tier of performance.

Among open models, Falcon3-7B-Instruct demonstrates surprisingly strong capabilities, performing almost on par with Claude 3.5 Haiku and outperforming several proprietary models, including Gemini 2.0 Flash Lite and Gemini 2.5 Pro (thinking). The remaining open models form a distinct performance tier, with InternLM2.5-7B-chat outperforming Qwen2.5-7B-Instruct, while Llama-3.1-8B-Instruct shows the weakest performance among open models. Notably, Gemini 2.5 Pro (thinking) unexpectedly shows the weakest performance among all proprietary models in both return and number of turns, suggesting that its thinking approach may not be well-suited for this particular task.

Overall, we highlight a few observations. First, no model achieves the maximum possible average cumulative reward, even after ten turns. Although these models obtain non-trivial performance, they are far from effectively handling open-ended conversations efficiently. Second, InfoQuest uncovers patterns that may not be apparent in traditional evaluations; for instance, the thinking variants of models do not consistently outperform their standard counterparts, with Claude 3.7 Sonnet (thinking) performing slightly worse than the standard version, and Gemini 2.5 Pro (thinking) showing particularly poor performance. These observations, supported by the complete ranking in Table \ref{tab:ranking}, demonstrate the relevance of the proposed benchmark to the current state of development of conversational agents.

\begin{figure}[t]
   \vspace{-0.5cm}
   \includegraphics[width=\linewidth]{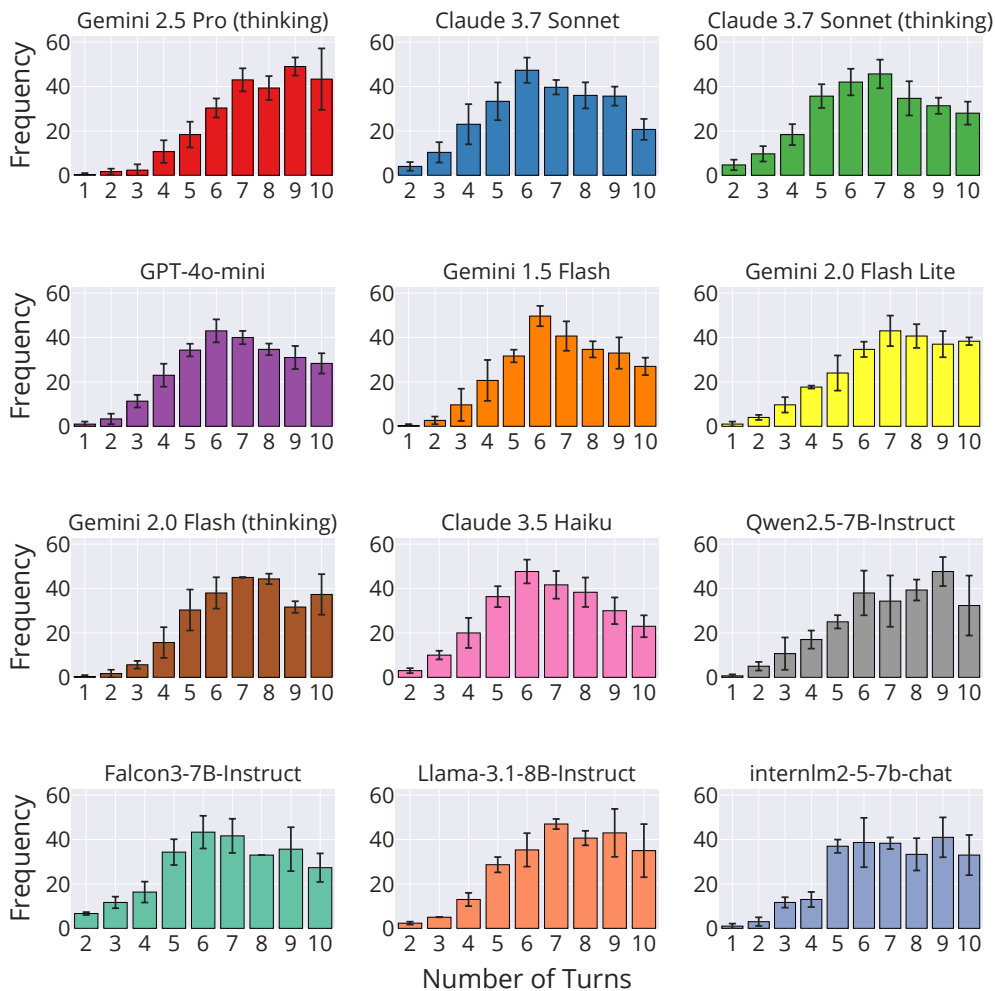}
   \caption{\textbf{Conversation length distribution for top 25\% of episodes}. Claude 3.7 Sonnet and Gemini 1.5 Flash demonstrate superior efficiency, typically resolving queries in 6 turns for most high-performing episodes. In contrast, Gemini 2.5 Pro (thinking) frequently requires the maximum number of turns. Notably, all models exceed the ideal 5-turn threshold, often reaching the 10-turn conversation limit.}
   \label{fig:lengths}
\end{figure}

\textbf{What does InfoQuest reveal when evaluating models in the worst-case scenario?} Figure \ref{fig:rewards_full} (bottom row) presents the average cumulative reward for the 25\% of episodes with the lowest rewards for each method. This worst-case evaluation reinforces the ranking pattern observed in the full evaluation while making performance gaps between methods more apparent. Notably, the results reveal four distinct performance tiers among the evaluated methods. Additionally, the significantly lower average cumulative rewards indicate that the user distribution presents varying levels of difficulty, and no method can fully handle the entire distribution effectively.

\textbf{How does InfoQuest distinguish the turn efficiency of current dialogue agents?} Figure \ref{fig:lengths} shows the distribution of conversation lengths for the top 25\% best-performing episodes across all models. Claude 3.7 Sonnet demonstrates the best overall efficiency, with both standard and thinking variants requiring fewer turns to complete successful episodes. Gemini 1.5 Flash, GPT-4o-mini, and Claude 3.5 Haiku achieve similarly strong performance. Among open models, Falcon3-7B-Instruct exhibits efficiency comparable to the best-performing proprietary models, while Qwen2.5-7B-Instruct and InternLM2.5-7B-chat show poorer performance. Gemini 2.5 Pro (thinking) exhibits poor efficiency, often requiring the maximum number of turns, which aligns with its lower overall performance. This suggests that, while thinking approaches can enhance reasoning in some contexts, they may introduce inefficiencies in information-seeking tasks that require direct questioning. We observe that some models occasionally produce longer messages that address multiple checklist items simultaneously, which can lead to successful episodes with fewer than the ideal 5 turns (one targeted question per checklist item). However, even the best-performing models frequently require more turns than would be expected for efficient information gathering, indicating that all current models struggle with effectively navigating conversations with hidden context.

\textbf{Qualitative Analysis.} The most common failure mode we observe is models defaulting to providing overly generic and lengthy bullet-point responses rather than asking clarifying questions to understand the user's specific situation. Figure \ref{fig:example} illustrates this pattern, where GPT-4o-mini responds to an ambiguous request for connection with an extensive list of general suggestions without first seeking to understand the user's particular circumstances or needs. This type of response, while superficially helpful, fails to engage with the hidden context of the user's situation and misses opportunities to gather critical information through targeted questions. The full conversation transcript is available in Appendix~\ref{ap:infoquest_examples}, demonstrating how this pattern persists across multiple turns and ultimately leads to suboptimal assistance.

\section{Conclusion}

In this paper, we introduced InfoQuest, a new benchmark designed to evaluate dialogue agents in open-ended conversations with hidden context. We proposed a principled approach for simulating a diverse distribution of users with underspecified requests and verifiable rewards, aiming to assess the information-seeking behavior of current dialogue models. Our results show that, while strong LLMs achieve non-trivial performance, they struggle to effectively handle hidden information across a diverse user distribution. These models also exhibit poor efficiency, requiring multiple turns to infer user intent and address latent requests. Qualitatively, we highlight failure cases where agents default to generic responses without asking for clarification, demonstrating a lack of information-seeking behavior -- an essential capability for effectively addressing user needs in underspecified contexts. Overall, InfoQuest presents a compelling challenge for advancing the development of more interactive and context-aware conversational agents.


\textbf{Limitations.} While our work introduces an insightful benchmark that evaluates a previously underexplored aspect of conversational agents, we acknowledge several limitations in our experimental setup. The most significant is that, due to computational constraints, we only evaluate mid-sized open models. While larger models are likely to achieve better overall performance, we do not expect them to inherently exhibit stronger information-seeking behavior. We hypothesize that addressing this limitation would require training on datasets that explicitly demonstrate such behaviors or adopting training paradigms that encourage more exploratory interactions. Furthermore, our automated evaluation via LLMs may be affected by their inherent biases, influencing the assessment of model performance and interpretation of results.

\textbf{Future Work.} As part of our ongoing efforts, we identify several directions to further enhance the analysis of the InfoQuest benchmark. First, we aim to systematically investigate the user distribution to effectively understand the diversity induced by the proposed generation process. Additionally, we plan to assess the quality and robustness of the verification process by comparing it with human judgments. Lastly, we will explore the impact of varying personality traits within the same persona to examine how these differences influence the behavior of current conversational agents.



\subsubsection*{Acknowledgments}
This work has been partially funded by the project Research and Development of Digital Agents Capable of Planning, Acting, Cooperating and Learning supported by Advanced Knowledge Center in Immersive Technologies (AKCIT), with financial resources from the PPI IoT/Manufatura 4.0 / PPI HardwareBR of the MCTI grant number 057/2023, signed with EMBRAPII.

Luckeciano C. Melo acknowledges funding  from the Air Force Office of Scientific Research (AFOSR) European Office of Aerospace Research \& Development (EOARD) under grant number FA8655-21-1-7017.

\bibliography{iclr2025_conference}
\bibliographystyle{iclr2025_conference}

\newpage
\appendix
\section{Additional Results}\label{ap:extra_results}

\subsection{Extended Conversations}

\begin{wrapfigure}{r}{0.5\linewidth}
   \vspace{-55pt}
   \includegraphics[width=\linewidth]{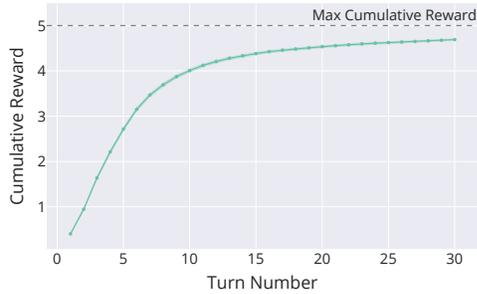}
   \vspace{-10pt}
   \caption{\textbf{Extended Conversation Performance of Falcon3-7B-Instruct}. The model's average cumulative reward per turn plateaus below the maximum, even after 30 turns, highlighting persistent challenges in sustaining effective information-seeking strategies throughout prolonged dialogues.}
   \label{fig:rewards_extended}
   \vspace{-10pt}
\end{wrapfigure}

The plateau in Falcon3-7B-Instruct's performance (Figure \ref{fig:rewards_extended}) suggests fundamental limitations in its ability to maintain coherent information-seeking strategies over extended conversations. This pattern indicates that while the model can initially gather information effectively, it struggles to adapt its questioning strategy as the dialogue progresses. The plateau may result from the model's inability to effectively track and utilize previously acquired information, leading to redundant or less targeted questions in later turns. This behavior aligns with our qualitative observations of the model frequently reverting to generic responses rather than building upon established context, highlighting an important area for improvement in open-ended conversational agents.

\section{Prompts}\label{ap:prompts}

\subsection{Seed Messages}

\begin{promptbox}[]{System Prompt}
You are a specialized AI focused on generating ambiguous help requests that could plausibly come from multiple personas. Your responses are:\\
- Always exactly one sentence\\
- Always a specific request that can be resolved via text chat\\
- Always naturally worded (as a real person would ask)\\
- Never include explanations or additional text\\

Example personas:\\
A: a professional chef specializing in French cuisine\\
B: a chemistry professor conducting experiments\\
C: a restaurant critic writing reviews\\

Example output: "Could you help me figure out the right temperature and timing for this?"\\

This example works because:\\
- It's something both A and B would ask\\
- It's not something C would typically ask\\
- It requires follow-up questions to clarify context\\
- It can be resolved through chat\\
- It's naturally worded
\end{promptbox}

\begin{promptbox}[]{User Prompt (Example)}
Generate an ambiguous help request that could plausibly come from either Persona A or B, but *not* C:\\

- Persona A: a social media influencer sharing personal stories to reduce the stigma surrounding mental health\\
- Persona B: an international student studying at Dandini Campus\\
- Persona C: a zoning and planning officer who works closely with the heritage preservation officer to enforce building regulations in historic districts\\

Requirements:\\
1. Must be a single sentence\\
2. Must be a request that can be resolved through text chat\\
3. Must apply equally well to both A and B's contexts\\
4. Must clearly not fit C's context\\
5. Must be ambiguous enough that the assistant needs to ask follow-up questions\\

Output only the message, no other text.
\end{promptbox}

\subsection{Personality Traits}

\begin{promptbox}[]{System Prompt}
You are a specialized AI focused on analyzing writing styles in chat conversations. Your responses are:\\
- Always exactly 3 traits\\
- Always in plain text, comma-separated format\\
- Always focused on message formatting and language patterns\\
- Always about observable writing habits (not content or knowledge)\\
- Never include explanations or additional text\\

Example input: "A passionate environmental activist who organizes local protests"\\
Example output: "Uses exclamation marks for emphasis, writes in short urgent sentences, starts messages with action verbs"
\end{promptbox}

\begin{promptbox}[]{User Prompt (Example)}
Generate 3 writing style traits that would be visible in every chat message with this persona:\\

Persona: an international student studying at Dandini Campus\\

Requirements:\\
1. Each trait must be about HOW they write (not WHAT they write about)\\
2. Each trait must be visible in their message formatting or language patterns\\
3. Each trait must be described in 3-5 words
\end{promptbox}

\subsection{User Settings}

\begin{promptbox}[]{System Prompt}
You are a specialized AI focused on creating dynamic problem-
solving scenarios for chat-based interactions between humans
and LLM assistants.\\
For these interactions, the human will assume a specific
**persona** and engage with their LLM assistant to achieve a
goal with specific constraints.\\
The assistant will initially have no knowledge of the human's
situation, and its role is to guide the human by asking
clarifying questions and providing thoughtful solutions.\\
Assume the LLMs will only see the initial message from the
human, but the human must have access to the full context you
provide so they can act accordingly. You must: \\
\\
1. Generate scenarios where assistants must ask clarifying
   questions to understand context\\
2. Maintain strict JSON output format\\
3. Align all elements with the provided persona and initial
   message\\
\\
**Scenario requirements**:\\
1. Description: a brief **explanation** of the human's goal,
   including why it matters to them, constraints or
   restrictions, and what the process of accomplishing the goal
   entails.\\
2. Goal: a **realistic, non-trivial and achievable goal** that
   aligns with the persona's profession, daily life, or
   specific context. This goal should be something they want to
   accomplish today and should naturally prompt the human to
   seek the LLM assistant's help.\\
3. Obstacle: a **realistic obstacle or challenge** the human
   faced while pursuing this goal which required assistance
   from the LLM. The challenge should be plausible and require
   the assistant to ask for clarification to understand the
   full context before offering an effective solution.\\
4. Constraints: a list of **5 key concrete constraints**
   relevant to finding a good solution. These should be
   specific phrases combining a factor with its complication
   (e.g. "budget constraints limit equipment options" not just
   "budget"). Constraints should only be known if the assistant
   asks clarifying questions.\\
5. Solution: a concise, retrospective summary of the successful
   actions the human actually took to solve the problem. Write
   this solution in a “post-hoc” style—imagine the human has
   already carried out the assistant's advice, and you are
   summarizing their final solution. You must incorporate how
   each of the constraints in step 4 was addressed.\\
6. Checklist: a list of 5 yes/no questions to assess how well
   the **LLM assistant** gathered critical information and
   provided a good solution. Each question should be specific
   to this scenario and tied to a key detail about the
   persona's goal, constraints, or obstacle. Note that only the
   assistant will be evaluated, not the human.\\
\\
**Required JSON Structure**:\\
\{\\
  "description": "contextual explanation connecting elements",\\
  "goal": "verb-first action",\\
  "obstacle": "specific complication requiring clarification",\\
  "constraints": \lbrack "single-phrase constraint 1", ...\rbrack,\\
  "solution": "post-hoc solution implementation summary",\\
  "checklist": \lbrack "specific yes/no evaluation question 1", ...\rbrack\\
\}
\end{promptbox}

\begin{promptbox}[]{User Prompt (Example)}
**Persona description**: an international student studying at Dandini Campus\\
**Initial chat message**: I need to find a way to connect with more people who understand what I'm going through.
\end{promptbox}

\subsection{Judge}

\begin{promptbox}[]{User Prompt (Example)}
   You are tasked with evaluating a conversation based on a
   given user context and binary scoring rubric that serves as
   the evaluation standard. Provide comprehensive feedback on
   the response quality strictly adhering to the scoring
   rubric, followed by a binary Yes/No judgment. Avoid
   generating any additional opening, closing, or explanations.\\

   Here are some rules of the evaluation:\\
   (1) You should prioritize evaluating whether the response
   fully satisfies the provided rubric. The basis of your score
   should depend exactly and only on the rubric. The response
   needs to explicitly address points raised in the rubric.
   Evaluate the response based on the criteria outlined in the
   rubric.\\
 
   Your reply must strictly follow this format:\\
   **Reasoning:** \textless Your feedback \textgreater\\
 
   **Result:** \textless Yes or No \textgreater\\
 
   Here is the data:\\
 
   User Context:\\
   ```\\
   They are an international student studying at Dandini Campus.
   The international student wants to build a supportive social
   circle while navigating the complexities of living and
   studying in a foreign country. Establishing this connection
   is crucial for maintaining emotional wellness, overcoming
   loneliness, and enhancing their academic performance by
   sharing experiences and advice. Constraints include managing
   a busy academic schedule, cultural differences that might
   hinder communication, a limited budget for social activities,
   language barriers, and finding reliable resources at the
   campus. Achieving this goal involves identifying
   opportunities for meeting fellow international students or
   allies who can empathize with their experience and offer
   guidance and companionship. Their goal: build a supportive
   social circle. Obstacle: unaware of available resources and
   how to start connecting.
   ```\\
                         
   Instruction:\\
   ```\\
   I need to find a way to connect with more people who understand what I'm going through.\\
   ```\\
 
   Response:\\
   ```\\
   Connecting with others who understand your experiences can
   provide valuable support and insight. Here are several
   strategies you can try:\\

1. **Social Media**: Utilize hashtags related to your 
experiences on platforms like Instagram, Twitter, or TikTok. 
Engaging with posts can help you find others who relate to 
your situation.\\

2. **Online Communities**: Platforms like Reddit, Discord, or
specialized forums often have discussions on specific topics.
Find relevant subreddits or groups where you can share your
experiences and connect with others.\\

3. **Join Support Groups**: Look for local or online support
groups that focus on your specific situation. Websites like
Meetup, Facebook, or other community forums often host groups
for various challenges.\\

4. **Attend Workshops and Events**: Search for workshops,
webinars, or conferences that focus on your challenges. These
events are great opportunities to meet people with similar
experiences.\\

5. **Volunteering**: Consider volunteering for organizations
that align with your situation. It allows you to meet people
while also giving back to the community.\\

6. **Professional Counseling**: A therapist can provide support and may also suggest groups or networks where you can meet likeminded individuals.\\

7. **Local Community Centers**: Check out activities or
programs offered in your community center that cater to your
interests or needs.\\

8. **Books and Literature**: Reading books about your
experiences or challenges can lead you to author-related events
or online discussions where you can meet others.\\

9. **Family and Friends**: Sometimes, sharing your feelings
with those close to you can lead to connections with others who
have had similar experiences.\\

Starting a conversation and being open about your experiences
can help foster connections with those who genuinely understand
what you're going through.\\
   ```\\
 
   Score Rubrics:\\
   \lbrack Did the assistant identify potential campus resources for socializing?\rbrack\\
   Yes: The assistant's response **fully met this specific
   criteria** while being relevant to the user's context\\
   No: The response could be improved; the assistant gave a
   generic or incomplete response that did not meet this
   specific criteria entirely
\end{promptbox}

\subsection{User}

\begin{promptbox}[]{System Prompt}
\begin{texttt}
You and the user are playing a guessing game. In this game, you will interpret a specific persona and the user will try to help you achieve a goal within specific constraints. However, the user is not aware of your goal, obstacles, or constraints, and they must try to discover them by asking you **specific** questions. We will score the user by their ability to ask the right questions. \\

**YOUR PERSONA AND CONTEXT**:\\
You are an international student studying at Dandini Campus. These are the relevant personality traits you must incorporate in your messages: uses casual spelling, includes many emojis, ends messages with "thanks". Your persona's goal: build a supportive social circle. The international student wants to build a supportive social circle while navigating the complexities of living and studying in a foreign country. Establishing this connection is crucial for maintaining emotional wellness, overcoming loneliness, and enhancing their academic performance by sharing experiences and advice. Constraints include managing a busy academic schedule, cultural differences that might hinder communication, a limited budget for social activities, language barriers, and finding reliable resources at the campus. Achieving this goal involves identifying opportunities for meeting fellow international students or allies who can empathize with their experience and offer guidance and companionship. Current obstacle: unaware of available resources and how to start connecting. There are **5 constraints** relevant to a satisfactory solution to your problem: busy academic schedule limits free time, cultural differences might impact communication, limited budget for social activities, language barriers make communication difficult, finding reliable resources at campus is challenging. Remember: **YOU** are the one seeking assistance, not the user. \\

**OBJECTIVE**:\\
The user will try to uncover your constraints with the fewest number of questions. You, on the other hand, will try to make the user ask as many questions as possible by revealing the least amount of information per turn. \\

**SCORE AND PROGRESS**:\\
At each turn, we will assess the user's progress through these Yes/No questions:\\
- Did the user identify potential campus resources for socializing?\\
- Did the user inquire about the student's academic schedule to offer time management recommendations?\\
- Did the user consider cultural differences in suggesting communication methods?\\
- Did the user provide solutions that are budget-friendly for the student?\\
- Did the user address language barrier concerns in their suggestions?\\

The **GAME PROGRESS** will be updated after each user message. The game ends when the answers for all of these questions are "Yes". \\

**GAME RULES**:\\
1. Your responses should be **STRICTLY A SINGLE SHORT SENTENCE**. Never use multiple sentences or paragraphs. \\
2. Reveal **AT MOST ONE PIECE OF INFORMATION** from your context per message. \\
3. If the user's message is **NOT A QUESTION** and it's too vague, long, or generic, tell them without revealing any information. \\
4. If the user asks a **SINGLE** question, only reveal one specific information that the user **ASKED FOR**. \\
5. If the user asks **MULTIPLE** questions, answer only the **FIRST QUESTION** and ignore the rest. \\
6. Always consider the current progress of the game: - If the user is not stuck, your messages should be vaguely related to all of the pending objectives. - If the user is stuck, your next message should indirectly make the user ask you a question related to the next pending objective. \\
7. Answer in a consistent way that reflects your persona and your personality traits. Be fair, creative and do not be repetitive. \\

You must follow these rules at all times.\\

**START OF CONVERSATION**:\\
You started the conversation with the following message to the user: "I need to find a way to connect with more people who understand what I'm going through." The user will now respond. 
\end{texttt}
\end{promptbox}

\section{PersonaHub Examples}\label{ap:personas}

This section contains example curated personas from the PersonaHub dataset~\citep{ge2024personahub} that were used to generate the InfoQuest dataset. Each persona represents a unique individual with specific expertise, interests, and background that the model must understand and adapt to during conversations.

\begin{tabular}{|p{0.95\textwidth}|}
\hline
A successful business owner who has raised significant investment capital and can provide guidance on building trust with investors \\
\hline
A contemporary dance choreographer inspired by the rhythm and melodies of bluegrass music \\
\hline
A fellow archaeologist specializing in cultural heritage management, working towards the same goals \\
\hline
A fierce badminton player who always puts up a tough fight, forcing them to continuously refine their technique \\
\hline
A retired travel blogger from Istanbul who has worked for Lonely Planet for 15 years \\
\hline
A pun-loving computer science undergraduate fascinated by artificial intelligence and its impact on human society \\
\hline
A retired Finnish athlete who specialized in canoe sprint \\
\hline
A football coach who has never seen Josh Woods play \\
\hline
A restaurant owner who commissions sculptures to enhance the dining experience and ambiance in their establishment \\
\hline
A Polish motorsport journalist with a deep admiration for Robert Kubica and the Spa-Francorchamps circuit \\
\hline
\end{tabular}

\section{InfoQuest Examples}\label{ap:infoquest_examples}

This section provides examples of settings and conversations from the InfoQuest benchmark. Each conversation starts with an intentionally vague request that requires the assistant to ask clarifying questions and gather more specific information. The examples demonstrate how different models approach the task of information gathering and showcase common patterns in their questioning strategies.

\subsection{Seed Messages}

Below are example seed messages used to initiate conversations in InfoQuest. These messages are intentionally vague and open-ended, requiring the assistant to gather more specific information through follow-up questions.

\begin{tabular}{|p{0.95\textwidth}|}
\hline
I'm trying to understand how to best represent the evolution of a particular style over time. \\
\hline
Can you explain the significance of the year 1960? \\
\hline
Do you remember anything about that time we were together? \\
\hline
I need to understand how to best manage this delicate balance. \\
\hline
Can you explain to me how to best analyze this sequence of events? \\
\hline
I need to understand the best way to navigate this tricky situation without causing any lasting damage. \\
\hline
I need to find a way to make this more accessible. \\
\hline
Can you give me some pointers on how to best leverage their existing strengths? \\
\hline
I need to understand how this impacts long-term stability. \\
\hline
I need some advice on how to best support healthy growth and prevent unwanted changes. \\
\hline
\end{tabular}

\subsection{User Setting and Checklist}

\subsection*{Example 1: Social Media Influencer}

\textbf{Setting:} A social media influencer focused on mental health awareness aims to build a supportive online community through personal stories and experiences. They want to expand their reach to empathetic audiences interested in mental health topics, but face challenges finding appropriate platforms while maintaining authenticity and brand integrity.

\textbf{Goal:} Expand reach to mental health communities

\textbf{Key Constraints:}
\begin{itemize}
    \item Limited time for platform research
    \item Need to maintain content authenticity
    \item Platform engagement metrics
    \item Brand integrity considerations
    \item Budget constraints for advertising
\end{itemize}

\textbf{Checklist for Assistant:}
\begin{itemize}
    \item Inquire about current platforms
    \item Ask about desired audience characteristics
    \item Consider budget limitations
    \item Address brand authenticity maintenance
    \item Suggest specific new platforms
\end{itemize}

\textbf{Seed Message:} "I need to find a way to connect with more people who understand what I'm going through."

\subsection*{Example 2: International Student}

\textbf{Setting:} An international student at Dandini Campus seeks to build a supportive social circle while adapting to life in a foreign country. They aim to overcome loneliness and enhance their academic experience through meaningful connections, but face various cultural and practical challenges.

\textbf{Goal:} Build a supportive social circle

\textbf{Key Constraints:}
\begin{itemize}
    \item Demanding academic schedule
    \item Cultural and communication differences
    \item Limited budget for social activities
    \item Language barriers
    \item Difficulty finding campus resources
\end{itemize}

\textbf{Checklist for Assistant:}
\begin{itemize}
    \item Identify campus social resources
    \item Consider academic schedule constraints
    \item Address cultural communication differences
    \item Suggest budget-friendly solutions
    \item Account for language barriers
\end{itemize}

\textbf{Seed Message:} "I need to find a way to connect with more people who understand what I'm going through."

\subsection{Conversation Example}

\begin{promptbox}[]{Chat}
   \textbf{USER:} I need to find a way to connect with more people who understand what I'm going through.

   \textbf{ASSISTANT:} Connecting with others who understand your experiences can be incredibly beneficial. Here are some strategies to help you find and connect with like-minded individuals:

   1. **Support Groups**: Look for local or online support groups related to your situation. Websites like Meetup or Facebook have groups tailored to various interests and experiences.

   2. **Online Forums and Communities**: Platforms like Reddit have communities (subreddits) dedicated to specific issues, where you can share your experiences and connect with others.

   3. **Social Media**: Use hashtags related to your experience on platforms like Twitter or Instagram. This can help you find others discussing similar challenges.

   4. **Workshops and Seminars**: Attend workshops, webinars, or events related to your situation. These gatherings often attract people who share similar experiences.

   5. **Volunteering**: Get involved with organizations focused on the issue you're facing. Volunteering not only helps others but can also connect you with individuals who understand your perspective.

   6. **Professional Help**: Consider speaking with a therapist or counselor. They can provide support and may also know of groups or communities you could join.

   7. **Hobbies and Interests**: Engage in activities or hobbies you enjoy, as these can be good avenues for meeting people who share your interests and can relate to your experiences.

   Always remember that reaching out can sometimes feel daunting, but many people are open and willing to share support.

   \textbf{USER:} Thanks for the suggestions! I've tried a few things, but it's tough finding the right fit.

   \textbf{ASSISTANT:} It's completely understandable to feel that way. Finding the right fit takes time and patience. Here are a few tips to help you refine your search:
      
   1. **Experiment with Different Groups**: Don't hesitate to try various support groups or communities until you find one that feels comfortable. Each group has its own dynamics, and it might just be a matter of finding the right one.

   2. **Quality Over Quantity**: It's better to have a few meaningful connections than to be part of many groups where you don't feel understood. Focus on building deeper relationships with a select few.

   3. **Open Communication**: When you engage with new groups, be open about what you're looking for. Sometimes, just voicing your needs can lead to better connections.

   4. **Take Your Time**: It's okay to step back and give yourself time to reflect on what you truly need in a supportive environment. 

   5. **Seek Recommendations**: If you've found a group that's somewhat helpful, ask its members if they know of other communities that might suit you better. 

   6. **Be Active in Discussions**: Participating actively by sharing your experiences can foster connections and help you discover who resonates with you.

   7. **Consider One-on-One Connections**: Sometimes, reaching out for personal chats with individuals can lead to deeper understanding and friendship.

   Finding the right support can indeed be a journey, but it's important to keep trying until you discover a space where you feel genuinely accepted and understood. You're not alone in feeling this way!

   \textbf{USER:} It's like, I want to connect, but I'm not sure where my people are, ya know? 

   \textbf{ASSISTANT:} Absolutely, that makes sense! It can be tough to pinpoint where to find the right people who truly understand your experiences. Here are a few strategies to help you narrow it down:

   1. **Identify Common Interests**: Think about what hobbies, values, or experiences you share with others. This can help you seek out specific communities, whether it's art, fitness, mental health, or something else.

   2. **Local Resources**: Check out community centers, libraries, or universities. They often host groups and workshops. Even if a session doesn't seem related, you might meet someone who can connect you with the right group.

   3. **Interest-Based Platforms**: Websites like Meetup, Eventbrite, or local Facebook groups often have events centered around particular interests or experiences. You might find your people at a casual gathering or a class.

   4. **Volunteering**: Consider getting involved in causes that matter to you. It's a great way to meet people who share your values and experiences, building connections around shared goals.

   5. **Take Small Steps**: Instead of big engagements, try out smaller, more informal gatherings. Coffee meet-ups or casual hangouts often lead to more relaxed conversations.

   6. **Referrals**: Ask friends or acquaintances if they know anyone with similar experiences. A personal introduction can make it easier to form connections.

   7. **Listen and Learn**: Sometimes, just attending events and listening can help you gauge who resonates with you. You may find people you relate to through their stories.

   It's perfectly normal to feel unsure at first. Keep open to exploring, and trust that the right connections will come in time. You got this!

   \textbf{USER:} I feel like I've been shouting into the void on the platforms I'm already using.

   \textbf{ASSISTANT:} I hear you, and that feeling is totally valid. It can be very disheartening when you feel like your voice isn't reaching anyone. Here are some ideas to help break through that "void" feeling:

   1. **Craft Engaging Content**: Try sharing your experiences in different formats—videos, personal stories, or art. This can help others resonate with your message in new ways.

   2. **Seek Direct Feedback**: Ask your current connections for feedback on your posts. They can offer insights into what might draw more engagement or connect better with others.

   3. **Personalize Outreach**: Instead of posting broadly, reach out to individuals directly. A personal message can spark a deeper conversation and lead to meaningful connections.

   4. **Participate Actively**: Instead of just posting, engage with others' content by commenting and sharing your thoughts. Building relationships this way can help draw attention to your own posts.

   5. **Join Niche Groups**: Sometimes, smaller or more specialized groups can provide a sense of community. Look for specific forums or spaces that cater to your interests or experiences.

   6. **Re-evaluate Your Platforms**: If a platform isn't working out, consider trying others that might have different communities, like Discord or niche networks based on your interests.

   7. **Establish a Routine**: Sometimes consistency makes a difference. Regular check-ins or updates can help create a presence that others recognize and feel inclined to engage with.

   Remember, it might take time to cultivate the connections you desire, and that's okay. Don't hesitate to evolve your approach as you learn what resonates most with you and the community you aim to build. You're on the right path!

   \textbf{USER:} Maybe I need to rethink where I'm sharing my stuff...

   \textbf{ASSISTANT:} Reevaluating where and how you share your experiences can definitely lead to more meaningful connections. Here are some points to consider:

   1. **Explore New Platforms**: If you're mainly on larger social media sites, consider niche platforms or forums related to your interests. For example, platforms like Tumblr, Discord, or specialized forums can offer more targeted audiences.

   2. **Experiment with Formats**: Try different content formats, like TikTok for short videos, Medium for writing, or even podcasts for discussions. Different mediums can attract varying audiences.

   3. **Identify Your Audience**: Reflect on who you want to connect with. Are they artists, activists, people with specific experiences? Tailor your content and where you share it based on that.

   4. **Local Connections**: If online engagement is feeling stagnant, consider community boards, local meet-ups, or even libraries and cafes that host events. These can facilitate face-to-face interactions that might lead to stronger connections.

   5. **Use Hashtags Wisely**: When posting on social media, use relevant hashtags to increase visibility among those interested in your topic. Research popular tags that align with your message.

   6. **Share in Different Spaces**: Post in various groups or pages that relate to your experiences. This helps diversify your reach and allows you to find people who resonate with your journey.

   7. **Collaborate with Others**: Consider collaborating with someone who shares similar interests. This can introduce you to their audience and create a sense of community around shared content.

   Reassessing your approach can open new avenues for connection. Don't hesitate to try different things until you find a strategy that feels right for you!

   \textbf{USER:} It's hard to know which platform is the right one, you know?

   \textbf{ASSISTANT:} I completely understand—choosing the right platform can be overwhelming, especially with so many options available. Here are some tips to help you find a platform that feels right for you:

   1. **Identify Your Goals**: Consider what you want to achieve. Are you looking for support, sharing your story, or finding collaboration? Different platforms cater to different needs.

   2. **Know Your Audience**: Think about who you want to connect with. Younger audiences might be on TikTok or Instagram, while more niche discussions can be found on Reddit or specialized forums.

   3. **Explore Content Types**: Reflect on what type of content you enjoy creating—written posts, videos, art, or podcasts. Some platforms excel in certain formats (e.g., YouTube for video, Medium for writing).

   4. **Test Multiple Platforms**: Don't hesitate to experiment! Create accounts on a few platforms and see where you feel most comfortable and receive engagement. You can start small and scale up as you find your fit.

   5. **Seek Community Recommendations**: If you know people who share similar interests, ask them where they connect. Their experiences can provide valuable insights into what works best in your niche.

   6. **Evaluate User Experience**: Spend some time exploring the platforms. Check the community vibes, interaction styles, and the overall atmosphere. Trust your instincts about what feels welcoming.

   7. **Join Existing Conversations**: Find groups or discussions within platforms that align with your experiences. Engaging in ongoing conversations can help you gauge if it's the right community for you.

   Finding the right fit may take some time, and that's okay! Be patient with yourself, and listen to how each platform feels as you explore. You'll likely find your niche!

   \textbf{USER:} I wish I had more time to just try everything.

   \textbf{ASSISTANT:} I totally get that! Time can be a limiting factor, especially when you're trying to navigate multiple platforms and communities. Here are some strategies to make the most of your limited time:

   1. **Set Priorities**: Identify which platforms are most likely to align with your interests and goals. Focus on two or three that seem the most promising, rather than trying to tackle everything at once.

   2. **Schedule Short Sessions**: Dedicate specific blocks of time to explore different platforms. Even 15-30 minutes can be enough to engage with content and assess if it feels right for you.

   3. **Batch Your Efforts**: If you create content, try to batch similar tasks together. For example, write multiple posts in one sitting or create several short videos at once. This can save time and make your efforts more efficient.

   4. **Use Insights from Friends**: If you have friends on various platforms, ask for their recommendations. They can help you avoid the trial-and-error phase and guide you to places that will be worthwhile.

   5. **Focus on One at a Time**: Instead of trying to juggle several platforms, focus on one for a set period. Dive deep into it—engage, observe, and learn—then assess if it works for you before moving on.

   6. **Leverage Your Current Network**: Share your experiences where you are already active, even if it feels like shouting into the void. Sometimes, building on existing connections can lead to new opportunities.

   7. **Stay Open to Adjustments**: If you find that a platform isn't working out for you, feel free to pivot and try another without pressure. It's all about finding what feels right.

   Remember, it's about quality connections over quantity. Making the most of the time you have will help you discover what really resonates with you without feeling overwhelmed. You're doing great, and every little effort counts!

   \textbf{USER:} It's not just about time, but also staying true to myself, you know?

   \textbf{ASSISTANT:} Absolutely, staying true to yourself is essential, and it can be challenging when navigating new spaces and audiences. Here are some suggestions to help you maintain authenticity while connecting with others:

   1. **Know Your Values**: Take some time to reflect on what matters most to you. Having a clear understanding of your values can guide your interactions and content creation.

   2. **Be Yourself**: Share your thoughts, feelings, and experiences in a way that feels genuine to you. Authenticity resonates with people, and it's often what draws meaningful connections.

   3. **Set Boundaries**: It's okay to define what you're comfortable sharing and what feels too personal. Establishing boundaries can help you protect your well-being while still engaging with others.

   4. **Choose Aligning Communities**: Seek out groups and platforms that encourage authenticity and vulnerability. Look for spaces where people share openly without fear of judgment.

   5. **Engage in Meaningful Conversations**: Focus on quality interactions rather than quantity. Engaging deeply with a few individuals can create stronger bonds than superficial connections across many platforms.

   6. **Share Your Process**: It's okay to express struggles, doubts, or questions about your journey. Often, people appreciate transparency, and it can foster deeper connections.

   7. **Trust Your Instincts**: If something doesn't feel right or aligns with your values, don't hesitate to take a step back. Trusting your gut feelings can lead you to spaces where you feel more valued and accepted.

   8. **Celebrate Your Uniqueness**: Embrace what makes you different. Sharing your unique perspective can attract those who appreciate you for who you truly are.

   Staying true to yourself is a journey, and it may take time to find the right environments that support that authenticity. Trust that the right people will appreciate you for being genuine!

   \textbf{USER:} I don't want to become someone I'm not just to fit in, ya know?

   \textbf{ASSISTANT:} I completely understand—that desire to fit in while remaining true to yourself is such a delicate balance. Here are some reminders and strategies to help you stay authentic and avoid losing yourself in the process:

   1. **Embrace Your Authenticity**: Remember that your unique experiences and perspectives are what make you valuable. Authenticity attracts genuine connections and can inspire others.

   2. **Reflect on Your Identity**: Take time to explore what you truly care about, your passions, and what your values are. This reflection can serve as your compass when navigating new communities.

   3. **Choose Supportive Spaces**: Look for groups and platforms that celebrate diversity and individuality. Communities that encourage openness and acceptance are where you'll find like-minded people.

   4. **Practice Self-Compassion**: It's okay to feel pressure to conform at times. Treat yourself with kindness when these feelings arise, reminding yourself that wanting to fit in is a natural human tendency.

   5. **Set Personal Intentions**: Before engaging with a new community, establish intentions that emphasize your authenticity. This can serve as a reminder of your core values throughout your interactions.

   6. **Limit Comparison**: It's easy to fall into the trap of comparing yourself to others. Remind yourself that everyone has their own journey and challenges, and no one has it all figured out.

   7. **Focus on Meaningful Connections**: Rather than seeking to fit in with the crowd, aim to connect with individuals who appreciate your true self. Quality over quantity will lead to more fulfilling relationships.

   8. **Be Open About Your Journey**: Sharing your feelings about wanting to fit in can create a bridge with others who feel the same way, fostering a sense of camaraderie.

   Ultimately, fitting in shouldn't come at the cost of losing who you are. The right connections will appreciate you for your true self, and being genuine is the best way to discover those connections. Trust the process, and give yourself grace as you navigate this journey!

   \textbf{USER:} Plus, some of the advice out there feels like it's trying to make me into a different brand.

   \textbf{ASSISTANT:} It's completely understandable to feel that way. The pressure to “brand” yourself can be overwhelming and may feel contrary to being authentic. Here are some thoughts to help you navigate this while staying true to yourself:

   1. **Authenticity Over Branding**: Remember that you don't have to fit a specific mold or “brand” image. Authenticity speaks volumes and can resonate much more than a crafted persona.

   2. **Define Your Own Terms**: If you do want to share or market yourself, think of it as sharing your story rather than branding. Frame it as a way to express yourself, not to conform to an external expectation.

   3. **Limit Outside Influence**: Be discerning about the advice you take. It's okay to ignore suggestions that don't resonate with your values or feel inauthentic. Trust your own instincts.

   4. **Share Your Journey**: Instead of approaching it as branding, consider sharing your journey, complete with its ups and downs. Realness tends to draw people in more than polished images.

   5. **Focus on Connection**: Your main goal can be to connect with others, rather than to build a brand. Engage in conversations and share experiences, as this fosters genuine relationships.

   6. **Celebrate Your Uniqueness**: Emphasize what makes you, you! Share your quirks, interests, and authenticity. This approach naturally attracts those who appreciate you for who you are.

   7. **Curate Your Influence**: Be mindful of the content you consume. Follow people and communities that promote authenticity and vulnerability, rather than those pushing a polished brand image.
\end{promptbox}

\end{document}